\documentclass[a4 paper, 11pt,twoside]{article}
\usepackage{fullpage}                           
\usepackage[lmargin=0.6in,rmargin=0.6in,tmargin=0.6in,headsep=.2in]{geometry}
\usepackage{graphicx}
\usepackage{forloop}
\usepackage{etoolbox}
\usepackage{calc}
\usepackage{wrapfig,lipsum,booktabs} 
\usepackage{xtab} 
\usepackage[dvipsnames]{xcolor}
\usepackage[normalem]{ulem}

\newcommand{\MyIL}{GMM-IL }
\newcommand{\MyILName}{GMM Incremental Learning }

\makeatletter
\def\@seccntformat#1{\@ifundefined{#1@cntformat}%
{\csname the#1\endcsname\;}
{\csname #1@cntformat\endcsname}
}
\def\section@cntformat{\thesection.\;} 
\def\subsection@cntformat{\thesubsection.\;} 
\makeatother
\usepackage{hyperref}
\hypersetup{
    colorlinks=true,
    urlcolor=blue,
    linkcolor=black,
    citecolor = black}

\usepackage{amsmath,amsfonts,amssymb,amsthm,epsfig,epstopdf,url,array}
\usepackage[retainorgcmds]{IEEEtrantools}
\usepackage{makeidx,epsfig,lscape}
\usepackage{xcolor,pict2e}
\usepackage{upgreek}

\theoremstyle{definition}

\begin{document}
\vspace*{0.5cm}
{\noindent\huge\bf GMM-IL: Image Classification using Incrementally Learnt,  Independent Probabilistic Models for Small Sample Sizes.}\\[1cm]
{\bf\large Penny Johnston, Keiller Nogueira, Kevin Swingler}\\[0.5cm]
Department of Computer Science, University of Stirling, Stirling, UK\\
Emails: penny.johnston, keiller.nogueira, kevin.swingler (@stir.ac.uk)\\

\raggedleft
{\color{Black}\rule{1\textwidth}{2pt}}
\raggedright
{\color{Black}\bf\large Abstract}\\
Current deep learning classifiers, carry out supervised learning and store class discriminatory information in a set of shared network weights. These weights cannot be easily altered to incrementally learn additional classes, since the classification weights all require retraining to prevent old class information from being lost and also require the previous training data to be present. We present a novel two stage architecture which couples visual feature learning with probabilistic models to represent each class in the form of a Gaussian Mixture Model. By using these independent class representations within our classifier, we outperform a benchmark of an equivalent network with a Softmax head, obtaining increased accuracy for sample sizes smaller than 12 and increased  weighted F1 score for 3 imbalanced class profiles in that sample range. When learning new classes our classifier exhibits no catastrophic forgetting issues and only requires the new classes' training images to be present. This enables a database of growing classes over time which can be visually indexed and reasoned over.
\vspace{0.5cm}\\
{\color{Black}\bf\large Keywords}\\
Image Classification, Incremental Learning, Probabilistic Models, Small Sample Sizes, Deep Learning
\vspace{0cm}\\
{\color{Black}\rule{1\textwidth}{2pt}}
\section{Introduction}
\label{sec:intro}
\begin{figure}[ht!]
\centering 
    \includegraphics[width=0.8\textwidth]{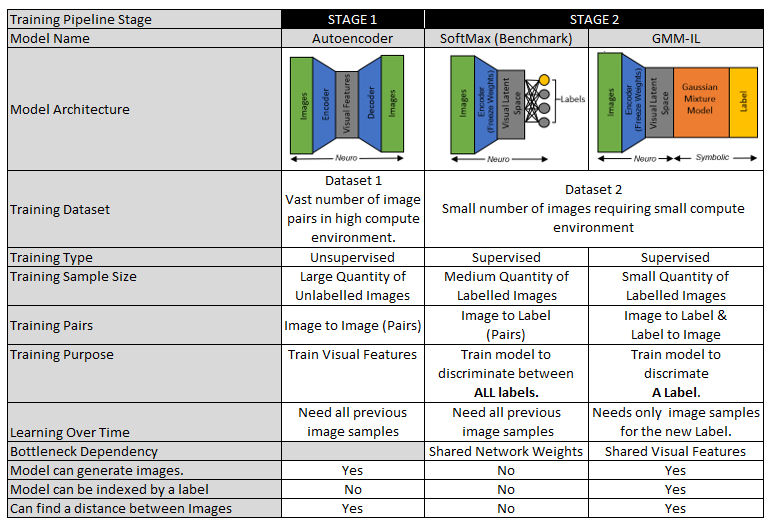}
    \caption{Two stage training process. Comparison of Autoencoder, GMM-IL and Benchmark Model Attributes.}
    \label{GMM Object}
\end{figure}
Incremental learning of new classes without forgetting old classes is essential for real-world problems but extremely challenging for modern deep learning methods. Current incremental deep learners suffer from `catastrophic forgetting' when after learning Class A, they are then required to learn Class B. The issue occurs due to the sharing of a set number of weights in the neural network. These are optimised for Class A, however, in order to learn Class B the weights must be altered, resulting in new knowledge overwriting previous knowledge. 

We take our inspiration from humans and create a two stage pipeline where images are first translated to visual features, which are then used during the modelling of independent classes. As shown in Figure \ref{GMM Object} the visual features are learnt from unlabelled data in stage one using an autoencoder, creating a latent space representation. Then in stage two each probabilistic model is independently trained to learn a class conditional probability distribution over the latent space. This enables new classes to be added without having to retrain the visual features or relearn previously learnt classes. The probabilistic model we use is a Gaussian Mixture Model. Using this architecture we can translate between label and image or image and label via the encoder and decoder created when training the autoencoder. The independent class models can be enriched with symbolic information and stored in an extensible knowledge graph. This proposed neuro-symbolic architecture creates a specific structure at four levels and each level is aligned with a human concept and supports open-ended learning which combines the strengths of the symbolic approaches with insights from machine learning. Figure \ref{fig:Overview} shows an overview of the proposed architecture, the analogies between humans and machines are shown below for each of the four levels: 
\begin{enumerate}
    \item Human: Learns to see visual information. 
    
    Machine: Learns a latent space of visual features during Autoencoder training. 
    \item Human: After seeing an object, are taught a name to give it meaning.
    
    Machine: After creating prior visual features, train a probabilistic model to represent a class which gives that group of training images a symbolic meaning. 
    \item Human: Learn a new object without needing to see previous objects at the same time. Previously learnt objects can be imagined.
    
    Machine: Train a new class in the form of a probabilistic model  without requiring access to previous training data. Previously learnt models can be sampled and the resulting latent embeddings decoded into images. 
    \item Human: Identify objects they are paying attention to in their field of view in real-time.
    
    Machine: Carry out object classification for the contents of an image at inference time.
\end{enumerate}

This paper is organised as follows: In Section \ref{Related_Work} we place our \MyILName within the ontology of Incremental Learners with associated discussion. We detail the proposed \MyIL in Section \ref{Proposed_Architecture}, document our experimental setup in Section \ref{Exp_Setup}, and subsequently report our results in Section \ref{Results}. In Section \ref{Discussion} we suggest possible improvements of this method. Finally, we conclude in Section \ref{Conclusions} and in Section \ref{Future_Research} we discuss what the proposed class representation enables at the boundary between images and labels in the form of disambiguation tasks.
\subsection{Contributions of this paper}
 To the best of our knowledge, there is no similar GMM-Incremental Learner in the literature. The value of our proposed method is in the delivery of :
\begin{itemize}
\item A novel class representation, which couples transferred visual feature learning with independent probabilistic class learning and is easily extended to accommodate new classes.
\item Gaussian Mixture Models, which can be trained using small sample sizes, decreasing both model training time and the need for costly annotated images.
\item A novel classifier that exhibits no catastrophic forgetting issues due to the removal of the shared weights found in the fully connected and softmax layers of standard deep learning classifiers.
\end{itemize}
\section{Related Work}
\label{Related_Work}
\subsection{Introduction}
Incremental Learning aims at incrementally updating a trained model, through tasks that learn new classes without forgetting old classes \cite{DeLange2022} \cite{Zhang} \cite{Kemker2018}. Class Incremental Learning is where a limited memory or no previously learned samples are allowed during the training process. This limitation is motivated by practical applications, such as storage and computing constraints which prevent us from simply retraining the entire model for each new task. It is worth mentioning, that Incremental Learning is different from Transfer Learning in that it also aims to have good performance in both old and new tasks. Since the objective of an Incremental Learner is to keep on learning new tasks, it should be evaluated based on the classifier's performance on the past and the present tasks, in order to be confident about its behaviour in future unseen tasks. Lopez-Paz and Ranzato \cite{Lopez-paz2017} pointed out that the ability of learners to transfer knowledge should also be paid attention to, and accordingly proposed the concepts of backward transfer (BWT, which is the influence that learning a task has on the performance of previous tasks) and forward transfer (FWT, which is the influence that learning a task has on the performance on future tasks). These new metrics are emerging which balance intransigence v forgetting \cite{Chaudhrya}.
\subsection{Incremental Learning Challenges}
Catastrophic Forgetting (CF) identified by McCloskey and Cohen \cite{McCloskey1989} over 30 years ago is when new learning interferes with the old learning, resulting in a reduced accuracy. Ideally, keeping the network’s weights stable prevents previously learned tasks from being forgotten, but too much stability prevents the model from learning new tasks. The essence of the stability-plasticity dilemma describes how to design a balanced system that is simultaneously sensitive to but not radically disrupted by new inputs and therefore can incrementally learn \cite{Kirkpatrick2017a}. 

\subsection{Ontology of Incremental Learners}
\begin{figure}[ht!]
\centering 
    \includegraphics[width=0.5\textwidth]{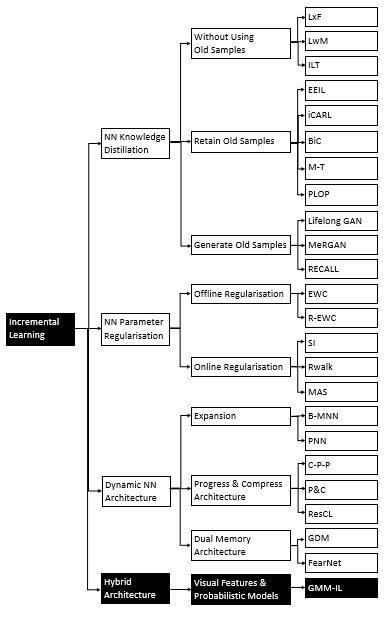}
    \caption{Ontology of Incremental Learners, placement of \MyIL in categories taken and adapted from Ritter et al \cite{Ritter}}
    \label{fig:sample_figure}
\end{figure}
The standard Incremental Learner models use a neural network framework, which intrinsically creates several challenges such as; catastrophic forgetting (\cite{Cohen2019}), memory limitation and concept drift. We adopt the ontology laid out in Ritter et al \cite{Ritter} selected due to it's structural categories. Other useful surveys and ontology's related to this field are \cite{Masana}, \cite{Kemker2018}, \cite{Parisi2019} , \cite{Luo2020} and \cite{Lange2021}. We reflect where \MyIL should be placed based on it's structure within this context. The three current categories specified to overcome Incremental Learning issues are:(1) Parameter Regularisation; (2) Knowledge Distillation; and (3) Dynamic Architecture.

Parameter Regularisation based methods utilise regularisation techniques such as constraining the update of important parameters, dropout and early stopping. These are all aimed at retaining previous task knowledge. Knowledge distillation methods distil knowledge from an old model into the current model. The various ways this is carried out are by; (1) retaining old samples \cite{Rebuffi}, (2) without using old samples and (3) generating old samples. With the addition of new tasks, most dynamic architecture methods flexibly adjust the network structure.

In our method, visual feature knowledge is represented in a static latent space, and the symbolic knowledge does not depend on any shared weights, so it does not require parameter regularisation. Whilst we do distill visual information into the latent space, this only needs to be done once during classifier initialisation. We do not need to manipulate or save old training samples in the form of exemplars during the incremental training of tasks. For these reasons \MyIL would belong in the Dynamic Architecture category.

Dynamic Architecture strategies include; (1) expansion, (2) progress and compress (P\& C) and (3) dual memory (D-M) architectures. Black-Modular neural networks (B-MNN) \cite{Terekhov}, Progressive neural networks (PNN)\cite{Liu2015} both augment the existing neural net with a 'piggy-back' neural net throughout the structure which gets trained on the new task. Not to be confused with fine tuning which adds one additional layer to a frozen memory. The main disadvantage of this approach is that the amount of parameters is exponentially proportional to the number of learned tasks. Progress and Compress (P\&C) architectures maintain a constant number of parameters and consist of two parts, a knowledge base and the active column. The compression phase extracts the knowledge learned in the previous expansion phase to the knowledge base, and uses the Elastic Weight Consolidation (EWC) \cite{Aich2021} strategy to protect the previously learned knowledge. In the expansion phase, the learning of new tasks reuses the characteristics of the knowledge base through lateral connections. The training approach alternates to limit expansion of the model while completing knowledge retention. These methods have limitations in scalability when it comes to multi-task incremental learning scenarios. Dual memory architectures are based on complementary learning systems (CLS) theory \cite{Mcclelland}, \cite{Kumaran2016}. The hippocampus system and the neocortex system balance the fast learning and slow learning processes. Therefore, the general dual memory architecture includes long and short-term memory. The former is used for memorizing past learning experiences and the later for learning current tasks. Growing Dual-Memory (GDM) considers the impact of continuous data over time on incremental learning.

Whilst we do expand our memory, each symbolic probabilistic model is small in size and expands at a rate of one model per learnt class. We do use a compress and expand strategy, but it fulfills a very different function. Our compress is for the visual features and our expand is in the form of building symbolic definitions. \MyIL does have a dual memory in the form of visual features held separately to symbolic definitions. However, we justify the creation of a new category called, 'Hybrid Architecture' by the fact that our proposed architecture does not solely use a neural net. This is created for future Incremental Learners that capture the benefits of neural nets and combine them with other models such as our probabilistic models. Following the trend within the ontology we add a structural sub category called, 'Visual Features \& Probabilistic Models', this reflects our separate visual and symbolic structure that delivers system stability and flexibility. We name our classifier ' \MyIL : \MyILName', which identifies our probabilistic model which is a Gaussian Mixture Model.
\section{Method}
\label{Proposed_Architecture}
Our aim is to classify a single visual concept using incremental learning, trained on small sample sizes using a hybrid architecture. The proposed architecture is modular in nature, enabling drop-in replacements for the Autoencoder and probability models. A description of the selection and training of these models can be found in Section \ref{Models}.

\begin{figure}[ht!]
\centering 
    \includegraphics[width=0.5\textwidth]{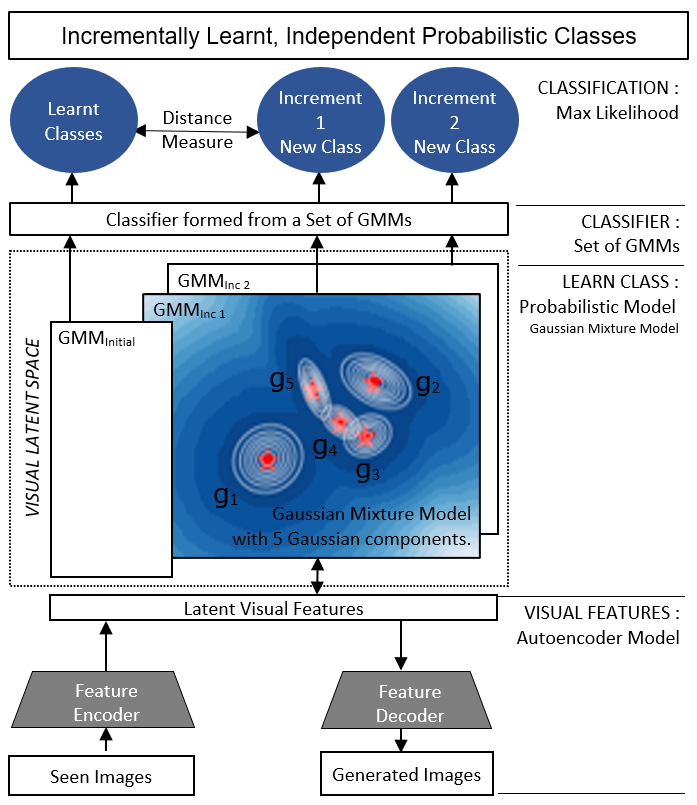}
    \caption{\MyIL: Incrementally learning Independent Probabilistic Class Models. (1) Autoencoder Model,(2) Probabilistic Model, (3) Classifier, (4) Classification Logic.
   \label{fig:Overview}}
\end{figure}
 The four levels and their interactions are shown in Figure \ref{fig:Overview} from the bottom up :
\begin{enumerate}
    \item An \textit{Autoencoder Model} trained once on a large corpus of unlabeled data, enabling generalised useful visual features to be extracted from the image corpus. Detailed in Section \ref{Stage1}.
    \item Independent \textit{Probabilistic Models} which form the class definition, independently trained on a small number of visual features. Visual features are a result of encoding the associated class images. Detailed in Section \ref{Stage2}.
    \item A \textit{Classifier} comprised of a set of learnt Probabilistic Models which can be added to as new class data become available. Detailed in Section \ref{Stage3}.
    \item \textit{Classification} logic can be carried out across all the probabilistic models to evaluate the likelihood that at inference time, a specific image belongs to a class. Detailed in Section \ref{Classification Logic}.
\end{enumerate}

\subsection{Autoencoder}
\label{Stage1}
The Autoencoder transforms an image from a high dimensional space to a lower dimensional space for ease of manipulation. The main premise is that unsupervised training initialises this encoder on a vast number of image samples in a high compute environment. When encoding visual features, we are not only interested in the autoencoder's ability to reconstruct the input image, but also on encoding a useful representation. By useful we mean the representation is not task specific, is spread throughout the latent space and contains visual motifs at different scales. These attributes will enable it to generalise well to unseen symbolic classes. Chadebec et al \cite{case} carried out a case study benchmark, where they presented and compared 19 generative autoencoder models. They found that the autoencoder which did not try to manipulate the latent space in end-to-end training produced the highest classification accuracy.

\subsection{Probabilistic Model}
\label{Stage2}
 The Probabilistic Model we selected is a Gaussian Mixture Model (GMM). This assumes all the data points are generated from a mixture of a finite number of Gaussian distributions. The GMM is created using the Maximum Likelihood Estimation (MLE) for normal mixtures using the Estimation Maximisation (EM) algorithm \cite{Dempster1977a}. A GMM is initialised using K-Means, trained with the following hyper-parameters:
\begin{enumerate}
    \item The number of mixture components to use.
    \item Covariance type: Each component has its own general covariance matrix, we use 4 types which are; Tied: all components share the same general covariance matrix. Diagonal: each component has its own diagonal covariance matrix. Spherical: each component has its own single variance. Full: each component has its own general covariance matrix.
    \item Non-negative regularisation: Added to the diagonal of the covariance. It ensures the covariance matrices are all positive. 
\end{enumerate}
The model with the lowest validation Bayesian Information Criterion (BIC) score is selected as the symbolic class representative.

\subsection{Classifier}
\label{Stage3}
Once the probabilistic model for one class has been learnt, we can incrementally learn the next one by simply training it and adding it to our set of GMMs in the classifier. This requires only the training data for the current class being learned and no previous classes' training data. Also, for the classifier to forget a class, it is as simple as removing the probabilistic class from the classifier set.

\begin{figure}[ht!]
\centering 
    \includegraphics[width=0.6\textwidth]{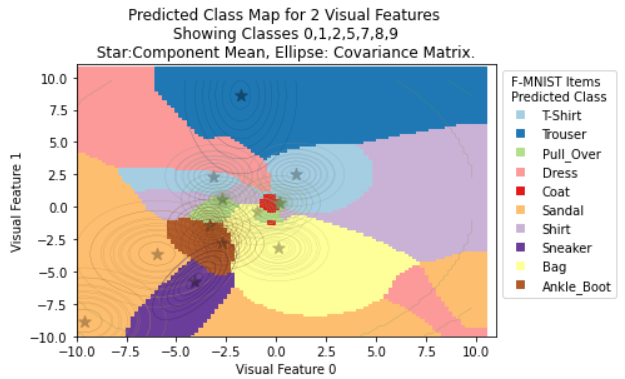}
    \caption{Map of the Predicted Classification for a 
 2 feature image encoder, overlaid with 7 GMMs and their component means (stars) and covariance values (ellipses) on the Fashion-MNIST dataset.
   \label{fig:Map}}
\end{figure}

In order to help with the intuition of a classifier comprised of a set of GMMs used for classification, we have built 10 GMMs based on a 2 feature encoder, this then enables us to create a 2D visualisation as shown in Fig \ref{fig:Map}. This map shows where individual GMM component distributions are, in the form of their GMM component mean values (stars) and co-variances (ellipses). We generate 2500 values for feature 1 \& 2 which represent encoded images, they cover our space and generate a map of what the predicted classification will be at each of these points, based on a maximum likelihood score. Only a few classes are shown so that the stars and ellipses are easier to see. When GMMs exist close together, the classifier has a reduced discriminatory power, classifier confusion, similar maximum likelihood scores and errors are likely where the Gaussian's meet.

\subsection{Classification Logic}
\label{Classification Logic}
At inference time we evaluate the likelihood of the image belonging to each GMM in the classifier and select the class with the maximum likelihood score as our classification.

\section{Experiment Setup}
\label{Exp_Setup}
\subsection{Hardware and software}
All deep learning-based models were implemented using TensorFlow \cite{Abadi2016}  Version 2.7.0. The code was written in Jupyter notebooks with Python Version 3.7.3. and CUDA version 11.2. All experiments conducted here were performed on a 64-bit Intel(R) Xeon(R) Gold 6130 CPU @ 2.10GHz workstation with  64 CPU cores and 768GB RAM. NVIDIA® GeForce (driver version 495.44) with 4*GTX1080Ti each with 11GB RAM. Debian version 10.12 was used as the operating system. The scikit-learn and pycm \cite{Haghighi2018} library were used for Metrics.
\subsection{Datasets}
The performance of our model and the benchmark model are evaluated on the public dataset Fashion-MNIST \cite{Vollgraf} which contains gray scale images of 10 clothing categories. The official training dataset was split into, 80\% creating a new training dataset (48K) and 20\% creating a new validation dataset (12K). 100\% of the official test dataset (10K) was used for our test set. Within each dataset all classes contained the same number of images, where this changes in our experiments it is noted in that experiments section. We use the same dataset for 2 purposes:
\begin{enumerate} 
    \item Dataset 1 - Autocoder Model Training : To train the Autoencoder model to create Visual Features. The premise is that the Autoencoder model will learn through unsupervised training on the largest image dataset possible, using vast computer power in a big data paradigm. That once carried out the resulting encoder/decoder will then be used on all vision tasks by without alteration (frozen weights). However, for the experiments in this paper we use the dataset above so that we can investigate if the classifier can learn unseen classes without that class having being used during the training of the Autoencoder. 
    \item Dataset 2 - Probabilistic Model Training: To train the GMMs to create independent symbolic class representations. It is this dataset that we manipulate to investigate the impact on the classifier accuracy of; sample size, imbalanced classes and incrementally learning class definitions.
\end{enumerate}
\subsection{Evaluation Metrics}
\label{Metrics}
We report the accuracy score to evaluate the classifier in the following experiments; baseline (Section \ref{Baseline}), Small Sample Sizes (Section \ref{Small Sample Sizes}) and Class Incremental Learning (Section \ref{Exp_Incremental_Learning}). We report the weighted F1 Score to evaluate the classifier in the, 'Imbalanced Class' section (Section \ref{Imbalanced_Class_Size}).
\subsubsection{Accuracy Score}
\label{Accuracy Score}
The accuracy score calculates the correctly predicted classes for a set of images.
\begin{equation}
	    Accuracy = \frac{(Number\;of\;Correctly\;Predicted\;Samples)}{Number\;of\;all\;Samples}
 \end{equation}
\subsubsection{Weighted F1 Score}
\label{F1_Score}
 The F1 score can be interpreted as a harmonic mean of the precision and recall, where an F1 score reaches its best value at 1 and worst score at 0. We weight the F1 Score so that the output average has accounted for the contribution of each class as weighted by the number of samples of that given class. The relative contribution of precision and recall to the F1 score are equal. The formula for the F1 score is: 
 \begin{equation}
	    F1_{Weighted} = 2 * \frac{(precision * recall)}{(precision + recall)} * \frac{(Class\;Sample\;Number)}{(Total\;Sample\;Number)}
\end{equation}
\subsection{Data Consistency}
\label{Data_Consistency}
When an experiment contains a suite of increasing or decreasing sample sizes, a dataset is managed to contain the same images as previously used to ensure experimental consistency. All reported test results are carried out using 100\% of the held out test dataset unless otherwise stated in an experiment.
\subsection{Model Setup}
\label{Models}
\subsubsection{Autoencoder}
\label{Feature_Encoding_Decoding}
\begin{figure}[ht!]
\centering 
    \includegraphics[width=0.7\textwidth]{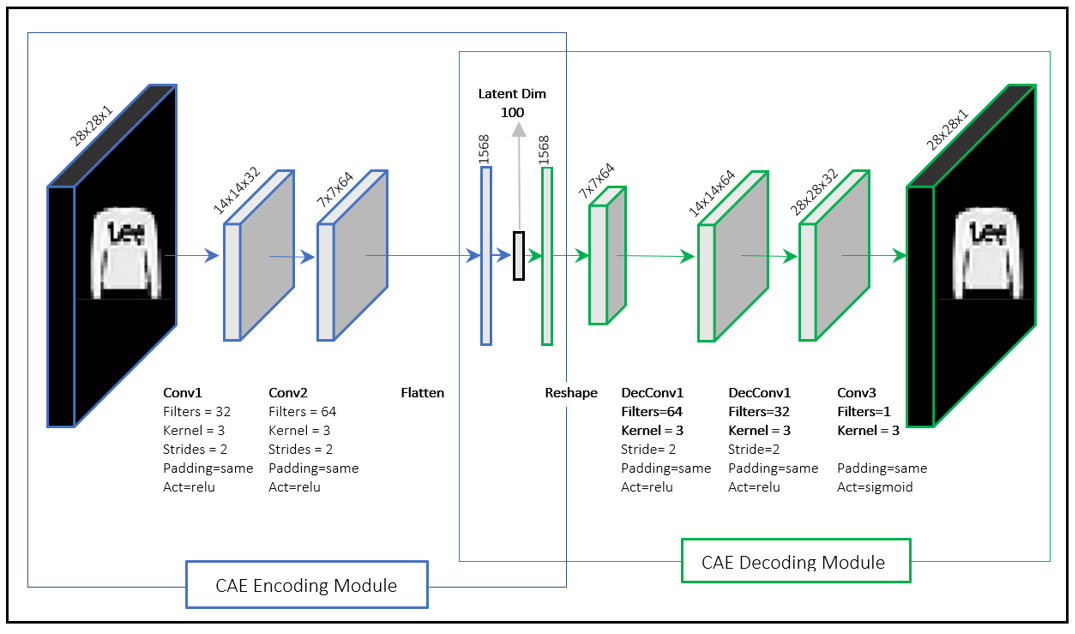}
    \caption{Autoencoder - A Convolutional Autoencoder containing the encoder to transform an image into a latent visual feature embedding with 100 features. Also the decoder which transforms a latent visual feature embedding back into an image.
   \label{fig:Encoder}}
\end{figure}

The encoder has two convolution layers (followed by ReLU activations \cite{Hinton} ) with 3 × 3 filters, applied with a stride of 2 and padding to maintain the same size image. From layer to layer, the number of filters (initially, 32) is doubled. The output of the last convolution layer is flattened and then mapped into a configurable dense layer which creates the features of our latent space. The decoder mirrors the encoder, using convolutional transpose operators \cite{Zeiler2010}. The full architecture is shown in Fig \ref{fig:Encoder}.

The autoencoder uses Adam \cite{Kingma2015} for optimisation. We reduced the learning rate according to a cosine function \cite{Loshchilov2017}, and through several experiments defined the search space for the following hyper-parameters, a base learning rate of 0.003, and a final learning rate of 0.001, a maximum number of 20 updates, 5 warm up steps and trained with 40 epochs. The batch size was set to 50. We train using unannotated images using a Mean Squared Error loss. 

We trained the Autoencoder as above, then holding all values the same except the latent dimension which we incrementally increased by steps of 10 features. We selected a size of 100 for the latent dimension since the larger the feature embedding the better the image reconstruction, measured by the minimum loss achieved. We required a feature representation that could be decoded into a good image representation whilst still compressing the data to enable easier manipulation. 

\subsubsection{Probabilistic Model}
\label{GMM}
The probabilistic model is a Gaussian Mixture Model (GMM) . A GMM was created for each symbolic class using encoded training images (100\% dataset unless otherwise stated in an experiment). We evaluated all the combinations of the following hyper parameters; i) Number of mixture components: 1 to 5 inclusive, ii) Covariance type: Tied, Diagonal, Spherical \& Full, and, iii) Non-negative regularization: 1.0e-2, 1.0e-3, 1.0e-4 and 1.0e-5. This resulted in 80 potential models for a class. During GMM model creation, occasionally when regularisation was low the maximum likelihood estimation (MLE) for normal mixtures didn't converge as the result of singularities or degeneracy. Any models which didn't converge were automatically eliminated from our potential selection. The selected GMM had the minimum validation BIC score.
\subsubsection{\MyIL Classifier (\textit{GMMs})}
\label{EnsembleOfGMMs}
We create our proposed classifier \textit{GMMs} by adding each learnt GMM to the set of GMMs. Table \ref{table:GMM_Config} shows the hyper parameters of our baseline set of GMMs. See Experiment \ref{Baseline} for this classifiers accuracy results.
\begin{table}[ht!]
\centering
\resizebox{0.5\textwidth}{!}{%
\begin{tabular}{|l|c|c|c|l|}
\hline
\textbf{Class} &
  \multicolumn{1}{l|}{\textbf{\begin{tabular}[c]{@{}l@{}}Reg\\ Covariance\end{tabular}}} &
  \multicolumn{1}{l|}{\textbf{\begin{tabular}[c]{@{}l@{}}Component\\ Number\end{tabular}}} &
  \multicolumn{1}{l|}{\textbf{\begin{tabular}[c]{@{}l@{}}Component \\ Shape\end{tabular}}} &
  \textbf{BIC} \\ \hline
0: T-Shirts   & 0.001 & 2 & full & 218599.8304 \\ \hline
1: Trousers   & 0.00001 & 2 & full & 92074.9725 \\ \hline
2: Pull Over  & 0.001 & 5 & tied & 215139.5721 \\ \hline
3: Dress      & 0.01 & 5 & tied & 218177.7049 \\ \hline
4: Coat       & 0.001 & 3 & tied & 209474.0943 \\ \hline
5: Sandal     & 0.0001 & 2 & full & 271939.6885 \\ \hline
6: Shirt      & 0.01 & 5 & tied & 225819.9335 \\ \hline
7: Sneaker    & 0.001 & 2 & tied & 166274.3903   \\ \hline
8: Bag        & 0.01  & 5 & tied & 293328.9243 \\ \hline
9: Ankle Boot & 0.01 & 3 & tied & 226256.1887 \\ \hline
\end{tabular}%
}
\caption{ GMM hyper parameters for set of GMMs in Baseline Classifier.}
\label{table:GMM_Config}
\end{table}

\subsubsection{Benchmark Classifier (\textit{Softmax})}
Our benchmark classifier (\textit{Softmax}) is comprised of a deep learning network consisting of the same frozen encoder model, plus a dense layer with a Softmax activation. We set the hyper parameters to the same as described for the Autoencoder (see Section \ref{Feature_Encoding_Decoding}). 

\section{Results and Analysis}
\label{Results}
These experiments investigate the difference in performance between a  multiple GMM head (\textit{GMMs}) and a benchmark method of a single Softmax head (\textit{Softmax}). Both classifiers use the same encoder with frozen weights trained on ten classes for all experiments except Experiment \ref{Exp_Incremental_Learning} where it was trained on six classes. The first Experiment \ref{Baseline} establishes a reference baseline. The next experiment evaluates the classifiers accuracy when using; small sample sizes during training  (Experiment \ref{Small Sample Sizes}) and when the sample size is imbalanced across classes (Experiment \ref{Imbalanced_Class_Size}). Experiment \ref{Exp_Incremental_Learning} reports the classifiers results when incrementally learning pairwise unseen classes. 
\subsection{Classifier's Baseline}
\label{Baseline}
We tested the two classifiers after building the models as described in Section \ref{EnsembleOfGMMs}. The results for training, validating and testing are shown in Table \ref{Table:Baseline}. \textit{Softmax} outperforms \textit{GMMs} when 100\% of each dataset is used and all classes are balanced.
\begin{table}[ht!]
\centering
\resizebox{0.5\linewidth}{!}{%
\begin{tabular}{|l|c|c|c|c|c|}
\hline
\textbf{Data} & 
\textbf{Classifier} &
\textbf{Acc} &
\textbf{F1} &
\textbf{CK} &
\textbf{MCC} \\ \hline
\begin{tabular}[c]{@{}l@{}}Train\end{tabular} &
\begin{tabular}[c]{@{}l@{}}GMM\end{tabular}
& 86.82\% & 86.32\% & 88.85\% & 85.44\% \\ \hline
\begin{tabular}[c]{@{}l@{}}Valid\end{tabular} &
\begin{tabular}[c]{@{}l@{}}GMM\end{tabular}
& 84.71\% & 84..25\% & 87.06\% & 83.22\% \\ \hline
\begin{tabular}[c]{@{}l@{}}Test\end{tabular} &
\begin{tabular}[c]{@{}l@{}}GMM\end{tabular}
& 85.57\% & 85.06\% & 87.15\% & 84.07\% \\ \hline
\begin{tabular}[c]{@{}l@{}}Train\end{tabular} &
\begin{tabular}[c]{@{}l@{}}Softmax\end{tabular}
& 97.97\% & 97.97\% & 98.23\% & 97.75\% \\ \hline
\begin{tabular}[c]{@{}l@{}}Valid\end{tabular} &
\begin{tabular}[c]{@{}l@{}}Softmax\end{tabular}
& 90.39\% & 90.39\% & 92.19\% & 89.33\% \\ \hline
\begin{tabular}[c]{@{}l@{}}Test\end{tabular} &
\begin{tabular}[c]{@{}l@{}}Softmax\end{tabular}
& 90.37\% & 90.35\% & 91.17\% & 89.31\% \\ \hline
\end{tabular}%
}
\caption{Classifier Accuracy for balanced classes using 100\% Training, Validation and Testing datasets. ACC: Accuracy, F1: Weighted F1 Score, CK: Cohen Kappa, MCC: Matthews Coefficient Correlation.}
\label{Table:Baseline}
\end{table}

\subsection{Small Sample Sizes}
\label{Small Sample Sizes}
\begin{figure}[h!]
\centering 
    \includegraphics[width=0.5\textwidth]{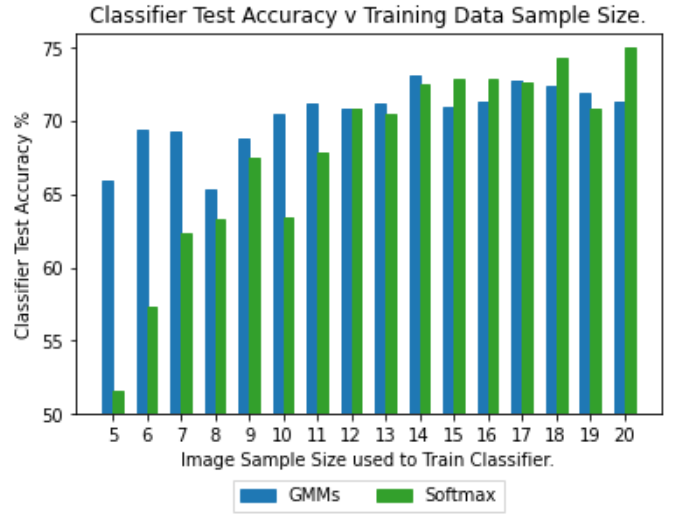}
    \caption{Classifier's test accuracy for a training sample size of 5 to 20 inclusive.
   \label{fig:Exp2}}
\end{figure}

Focusing on small sample sizes we selected a range of 5 to 20 (inclusive) samples. Stepping through each sample size we retrained all GMM models for both classifiers using the initial hyper parameter settings.  As can be seen in Figure \ref{fig:Exp2}. \textit{GMMs} perform with higher accuracy than \textit{Softmax} for sample sizes smaller than 12. 

\subsection{Imbalanced classes}
\label{Imbalanced_Class_Size}
\begin{figure}[ht!]
\centering 
    \includegraphics[width=0.5\textwidth]{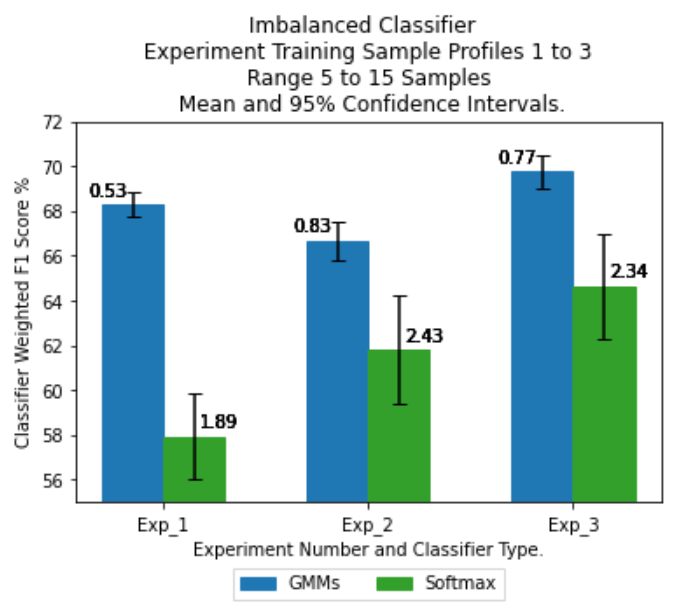}
    \caption{Classifier Weighted F1 Score for 3 Imbalanced Training Dataset Profiles. (Exp\_1 : 1 class n15 \& 9 classes n5), (Exp\_2 : 5 classes n15 \& 5 classes n15), (Exp\_3 : Classes start at 5 samples and increment to 14 samples.), Classes rotated 10 times. Mean and 95\% confidence intervals shown.
   \label{Fig:Exp_5.31}}
\end{figure}

In the classification problem field, the scenario of imbalanced classes \cite{Krawczyk2016} appears when the numbers of samples that represent the different classes are very different. The minority classes are usually the most important concepts to be learnt, since they represent rare cases or because the data acquisition of these examples is costly. In this work we create three imbalanced ratio profiles and report the classifiers weighted F1 Score (Section \ref{F1_Score}). Focusing on small data sizes we select the range 5 to 15 which was shown in the 'Small Sample Sizes' experiment as been of interest. Using 5 as low and 15 as high we created 3 imbalanced class datasets. We create imbalances that covered; (1) Extreme ratio difference of 1 class high and 9 classes low. (2) A 50:50 ratio difference of 5 class high, 5 class low and, (3) A stepped profile, Classes start at 5 samples and increment 1 sample until 14 samples. Both classifiers were retrained using the initial hyper parameters, GMM models where then fitted to new sample sizes based on the imbalanced experiment profile. Each Experiment was repeated 10 times with the class numbers rotating through the experiment profile. Figure \ref{Fig:Exp_5.31} shows the mean accuracy and 95\% confidence intervals per experiment and classifier type. Experiment 1,2 and 3 had p values of 0.000, 0.001 and 0.018 respectively. In all three experiments the \textit(GMM) out performed the \textit(Softmax) when trained on sample sizes under 15. 

\subsection{Class Incremental Learning}
\label{Exp_Incremental_Learning}
\textit{Softmax} classifiers learn all classes at once using all the training data. They do not perform as accurately when they are required to learn classes over time and have no access to previous training data.

\begin{figure}[ht!]
\centering 
    \includegraphics[width=0.5\textwidth]{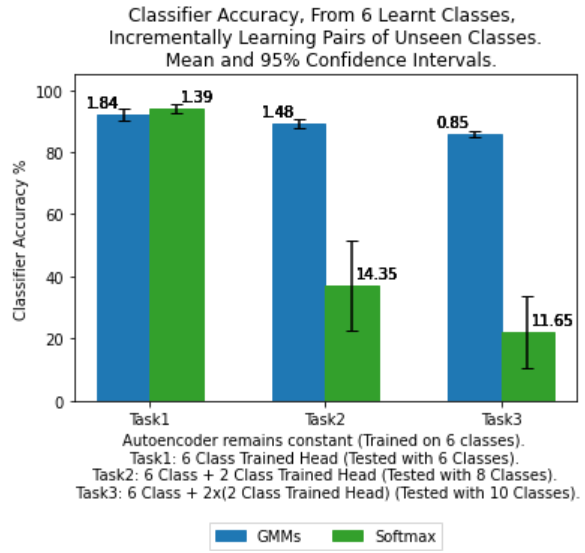}
    \caption{Classifiers Incrementally Learning Three Tasks.
   \label{Incremental_Learning}}
\end{figure}

We follow the benchmark method taken from Krichmar et al \cite{Kolouri2019}, and DeepMind, Google Research \cite{Kirichenko2021} who used the Split MNIST  dataset to learn consecutive pairs. For our dataset this is pairs of clothes eg., Pair 1: T-Shirt,Trousers, Pair 2: PullOver, Dress, Pair 3: Coat, Sandal, Pair 4: Shirt, Sneaker, Pair 5: Bag, Ankle Boot. We then make the following adjustments. We combine the first 3 pairs, which makes 6 classes trained in Task 1. We first train the Autoencoder model using the 6 classes and freeze the resulting encoder. Then we train the 6 GMM Models using their encoded images. This frozen encoder is then used for further tasks with just the GMM Models been trained. We use Pair 4 for Task 2 and Pair 5 for Task 3. The reason we create Tasks which contain 2 classes is to enable the \textit{Softmax} to classify without having access to prior training data. For clarification the 3 Tasks were configured as follows: 
\begin{enumerate}
    \item Task 1 established the accuracy when the encoder was trained on 6 classes, the classifier heads (\textit{GMM} and \textit{Softmax}) were tested on 6 classes using 100\% datasets. The classification was assigned to the class with the greatest probability/likelihood.
    \item Task2 established the accuracy when the classifiers were trained as per Task1 with 2 further classes, the classifier heads (\textit{GMM} and \textit{Softmax}) were tested on 8 classes using 100\% datasets. The classification was assigned to the class with the greatest probability/likelihood.
    \item Task3 established the accuracy when the classifiers were trained as per Task2 with 2 further classes, the classifier heads (\textit{GMM} and \textit{Softmax}) were tested on all 10 classes using 100\% datasets. The classification was assigned to the class with the greatest probability/likelihood.
\end{enumerate}
Task1, Task2 and Task3 were repeated 10 times as the classes were rotated, the mean and 95\% confidence values were calculated across all 10 combinations per classifier type.
From the results shown in Fig \ref{Incremental_Learning} it can be seen that initially the \textit{Softmax} is more accurate than the \textit{GMMs}. However, after each Incremental Task, the \textit{Softmax} accuracy decreases significantly more than \textit{GMM}. This shows the \textit{GMMs} has a greater ability to retain class definitions than the \textit{Softmax}.

\section{Discussion}
\label{Discussion}
\subsection{Introduction}
We created an architecture which enables transferred visual learning and incremental addition of class definitions in the form of probabilistic models. The visual learning we carried out used a smaller sample than would ultimately be used to enable us to control the content of the visual features and verify that unseen classes could be learnt. 
\subsection{Classifier Improvements}
This classifier's accuracy could be improved by making the following amendments.
\begin{itemize}
    \item Autoencoder - Our main premise is that this model's accuracy is dependent on the quality of the latent space created by the autoencoder, by using a state of the art autoencoder the granularity and quality of the visual features will be improved and hence the discriminatory power of the classifier increased.
    \item Gaussian Mixture Models - Amendola et al \cite{Amendola2020} state that there is the possibility of more modes than means when Gaussians are combined. Further investigation needs to be carried out to optimise the accuracy of the GMM likelihood landscape for a set of GMMs. 
\end{itemize}
\subsection{GMM as a Generative Model}
\begin{figure}[h!]
\centering 
    \includegraphics[width=0.5\textwidth]{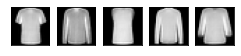}
    \caption{Reconstructed images generated from the Shirt Gaussian Mixture Model's component's mean points in the Fashion-MNIST dataset.
   \label{fig:Exp6}}
\end{figure}

The GMMs are a type of generative model. This means that it can be sampled to obtain it's associated visual features. The visual features in turn can be decoded into an image. This gives us the visual ability to literally see the concept that a GMM has modelled. A task designer or domain expert can intuitively check that the visual representation is aligned with the intended definition, delivering explainable class definitions and the ability to create synthetic data.  Figure \ref{fig:Exp6}. shows the images generated at the GMMs component means for the Shirt class found in the Fashion-MNIST dataset.
\section{Conclusions}
\label{Conclusions}
In conclusion, we propose a useful class representation where we learn visual features using unsupervised training. Using these we train independent probabilistic class definitions which incorporate uncertainty. We use these representations within a classifier and benchmark it to an equivalent Softmax classifier which uses class relative probability. We find \MyIL is more accurate for sample sizes smaller than 12 images and more robust for three imbalanced class profiles in the same sample range. \MyIL could incrementally learn the class definitions with no catastrophic forgetting issues which the Softmax benchmark exhibited. We conclude that in a learning environment where only small sample sizes are available and no previous training data is available this model shows potential as a classifier.
\section{Future Research}
\label{Future_Research}
\subsection{Introduction}
In this paper the classification logic we implement is to predict the class by calculating the maximum likelihood class. However, the creation of these independent, Gaussian Mixture Model Classes, enables further downstream functionality.
\subsection{Perceptual Reasoning}
The GMM definitions can be accessed using an index and used by a higher level logic with access to additional contextual information. A pairwise GMM distance can be calculated which gives an indication of how similar the GMM representations are, and hence the extent of the classifiers discriminatory power. We tested this by creating a pairwise distance matrix for the Fashion-MNIST dataset, using the Jensen-Shannon method, then correlated it with the confusion matrix created at inference time. The resulting Spearman correlation coefficient was 0.78166 with a p value of 0, which indicated that the similarity matrix could predict the level of classifier errors to some extent.

 The independent structure of the GMMs is ideal to underpin a multi class, multi instance environment. If we were to use the GMMs as nodes in a knowledge graph and the pairwise distances as edge weightings, we can create a static directed graph to reason over. Two possible scenarios when the presence of additional contextual information would help disambiguate are: 
\begin{enumerate}
    \item Illusions: In a bottom up scenario when an agent finds 2 labels for the same image (pairwise distance is small) then with the addition of further information such as the physical location the correct classification could be found.
    \item Homophones: In a top down scenario when a Natural Language Processing (NLP) task hears a word which could have 2 meanings, additional image capture could be carried out to see if there is an object present which clarifies the meaning.
\end{enumerate}

 This paper describes the creation of symbolic definitions for items, this could be expanded to define descriptive adjectives, normal verbs and affordances. Which could also be used to aid reasoning in a Knowledge Graph.

\subsection{Continual and Lifelong learning}
The model has the potential to be extended into the realms of Continual and Lifelong learning. Adding the functionality to manipulate the GMM definition as an agent interacts with an environment would enable real time symbolic perceptual adaptation and grounding ability. Research is needed to investigate when the GMM definition needed updating or at what point a new class should be created and labelled. 

\subsection{Federated Learning}
These probabilistic models and their encoders/decoders would be useful in the federated learning setting, where the premise is that the models move to meet the data, instead of the data moving to meet the models. As new classes can be added to a classifier without reference to previous training data, a federated network could allow multiple users to submit new classes without the need for access to data belonging to others. Classifiers composed of probabilistic models from different datasets could enable blended learning by selecting specific class definitions (GMMs) to create a turnkey classifier.
\newpage
\section*{Acknowledgement}
\addcontentsline{toc}{section}{Acknowledgement}
We would like to thank Stirling University for funding this research.
\def\cprime{$'$} \def\cprime{$'$}
{\color{Black}
\addcontentsline{toc}{section}{References}
}
{\color{black}

\expandafter\ifx\csname natexlab\endcsname\relax\def\natexlab#1{#1}\fi
\providecommand{\bibinfo}[2]{#2}
\ifx\xfnm\relax \def\xfnm[#1]{\unskip,\space#1}\fi
    \bibliographystyle{unsrt}
    \bibliography{biblist}

\begin{thebibliography}{10}

\bibitem{DeLange2022}
Matthias {De Lange}, Rahaf Aljundi, Marc Masana, Sarah Parisot, Xu~Jia, Ales
  Leonardis, Gregory Slabaugh, and Tinne Tuytelaars.
\newblock {A Continual Learning Survey: Defying Forgetting in Classification
  Tasks}.
\newblock {\em IEEE Transactions on Pattern Analysis and Machine Intelligence},
  44(7):3366--3385, 2022.

\bibitem{Zhang}
Song Zhang, Computer Science, Gehui Shen, Computer Science, Jinsong Huang,
  Computer Science, Zhi-hong Deng, and Computer Science.
\newblock {Self-Supervised Learning Aided Class-Incremental Lifelong Learning}.
\newblock pages 1--12.

\bibitem{Kemker2018}
Ronald Kemker, Marc McClure, Angelina Abitino, Tyler~L. Hayes, and Christopher
  Kanan.
\newblock {Measuring catastrophic forgetting in neural networks}.
\newblock {\em 32nd AAAI Conference on Artificial Intelligence, AAAI 2018},
  pages 3390--3398, 2018.

\bibitem{Lopez-paz2017}
David Lopez-paz and Marc~Aurelio Ranzato.
\newblock {Gradient Episodic Memory for Continual Learning}.
\newblock (Nips), 2017.

\bibitem{Chaudhrya}
Arslan Chaudhry, Puneet~K Dokania, Thalaiyasingam Ajanthan, and Philip H~S
  Torr.
\newblock {arXiv : 1801 . 10112v2 [ cs . CV ] 19 Mar 2018 Understanding
  Forgetting and Intransigence}.
\newblock (1).

\bibitem{McCloskey1989}
Michael McCloskey and Neal~J. Cohen.
\newblock {Catastrophic Interference in Connectionist Networks: The Sequential
  Learning Problem}.
\newblock {\em Psychology of Learning and Motivation - Advances in Research and
  Theory}, 24(C):109--165, 1989.

\bibitem{Kirkpatrick2017a}
James Kirkpatrick, Razvan Pascanu, Neil Rabinowitz, Joel Veness, Guillaume
  Desjardins, Andrei~A. Rusu, Kieran Milan, John Quan, Tiago Ramalho, Agnieszka
  Grabska-Barwinska, Demis Hassabis, Claudia Clopath, Dharshan Kumaran, and
  Raia Hadsell.
\newblock {Overcoming catastrophic forgetting in neural networks}.
\newblock {\em Proceedings of the National Academy of Sciences of the United
  States of America}, 114(13):3521--3526, 2017.

\bibitem{Ritter}
Hippolyt Ritter and David Barber.
\newblock {Online Structured Laplace Approximations For Overcoming Catastrophic
  Forgetting}.

\bibitem{Cohen2019}
Taco~S. Cohen, Mario Geiger, and Maurice Weiler.
\newblock {A general theory of equivariant CNNs on homogeneous spaces}.
\newblock {\em Advances in Neural Information Processing Systems}, 32(NeurIPS),
  2019.

\bibitem{Masana}
Marc Masana, Xialei Liu, Bart{\l}omiej Twardowski, Mikel Menta, Andrew~D
  Bagdanov, Joost Van~De Weijer, and L~G May.
\newblock {Class-incremental learning : survey and performance evaluation on
  image classification}.
\newblock pages 1--26.

\bibitem{Parisi2019}
German~I Parisi, Ronald Kemker, Jose~L Part, Christopher Kanan, and Stefan
  Wermter.
\newblock {Continual lifelong learning with neural networks : A review}.
\newblock {\em Neural Networks}, 113:54--71, 2019.

\bibitem{Luo2020}
Yong Luo, Liancheng Yin, Wenchao Bai, and Keming Mao.
\newblock {An appraisal of incremental learning methods}.
\newblock {\em Entropy}, 22(11):1--27, 2020.

\bibitem{Lange2021}
Matthias~De Lange, Rahaf Aljundi, Marc Masana, Sarah Parisot, and Xu~Jia.
\newblock {A continual learning survey : Defying forgetting in classification
  tasks}.
\newblock 8828(c):1--20, 2021.

\bibitem{Rebuffi}
Sylvestre-alvise Rebuffi and Kai Han.
\newblock {LSD-C : Linearly Separable Deep Clusters}.

\bibitem{Terekhov}
Alexander~V Terekhov, Guglielmo Montone, and J~Kevin~O Regan.
\newblock neural networks.

\bibitem{Liu2015}
Xiaodong Liu.
\newblock {Representation Learning Using Multi-Task Deep Neural Networks for
  Semantic Classification and Information Retrieval}.
\newblock pages 912--921, 2015.

\bibitem{Aich2021}
Abhishek Aich.
\newblock {Elastic Weight Consolidation (EWC): Nuts and Bolts}.
\newblock pages 1--7, 2021.

\bibitem{Mcclelland}
James~L Mcclelland, Bruce~L Mcnaughton, and Randall C~O Reilly.
\newblock {Why there are Complementary Learning Systems in the Hippocampus and
  Neocortex : Insights from the Successes and Failures of Connectionist Models
  of Learning and Memory}.

\bibitem{Kumaran2016}
Dharshan Kumaran, Demis Hassabis, and James~L Mcclelland.
\newblock {What Learning Systems do Intelligent Agents Need ? Complementary
  Learning Systems Theory Updated}.
\newblock {\em Trends in Cognitive Sciences}, 20(7):512--534, 2016.

\bibitem{case}
A~Benchmarking~Use Case, Cl{\'{e}}ment Chadebec, and Louis~J Vincent.
\newblock {Pythae : Unifying Generative Autoencoders in Python}.
\newblock pages 1--50.

\bibitem{Dempster1977a}
A.~P. Dempster, N.~M. Laird, and D.~B. Rubin.
\newblock { Maximum Likelihood from Incomplete Data Via the EM Algorithm }.
\newblock {\em Journal of the Royal Statistical Society: Series B
  (Methodological)}, 39(1):1--22, 1977.

\bibitem{Abadi2016}
Mart{\'{i}}n Abadi, Paul Barham, Jianmin Chen, Zhifeng Chen, Andy Davis,
  Jeffrey Dean, Matthieu Devin, Sanjay Ghemawat, Geoffrey Irving, Michael
  Isard, Manjunath Kudlur, Josh Levenberg, Rajat Monga, Sherry Moore, Derek~G
  Murray, Benoit Steiner, Paul Tucker, Vijay Vasudevan, Pete Warden, Martin
  Wicke, Yuan Yu, Xiaoqiang Zheng, Google Brain, Implementation Osdi, Paul
  Barham, Jianmin Chen, Zhifeng Chen, Andy Davis, Jeffrey Dean, Matthieu Devin,
  Sanjay Ghemawat, Geoffrey Irving, Michael Isard, Manjunath Kudlur, Josh
  Levenberg, Rajat Monga, Sherry Moore, Derek~G Murray, Benoit Steiner, Paul
  Tucker, Vijay Vasudevan, Pete Warden, Martin Wicke, Yuan Yu, and Xiaoqiang
  Zheng.
\newblock {TensorFlow : A System for Large-Scale Machine Learning This paper is
  included in the Proceedings of the TensorFlow : A system for large-scale
  machine learning}.
\newblock 2016.

\bibitem{Haghighi2018}
Sepand Haghighi, Masoomeh Jasemi, and Shaahin Hessabi.
\newblock {PyCM : Multiclass confusion matrix library in Python}.
\newblock 6(4):2--3, 2018.

\bibitem{Vollgraf}
Roland Vollgraf.
\newblock {Fashion-MNIST : a Novel Image Dataset for Benchmarking Machine
  Learning Algorithms}.
\newblock pages 1--6.

\bibitem{Hinton}
Geoffrey~E Hinton.
\newblock {Rectified Linear Units Improve Restricted Boltzmann Machines}.
\newblock (3).

\bibitem{Zeiler2010}
Matthew~D Zeiler, Dilip Krishnan, Graham~W Taylor, and Rob Fergus.
\newblock {Deconvolutional Networks}.
\newblock pages 2528--2535, 2010.

\bibitem{Kingma2015}
Diederik~P. Kingma and Jimmy~Lei Ba.
\newblock {Adam: A method for stochastic optimization}.
\newblock {\em 3rd International Conference on Learning Representations, ICLR
  2015 - Conference Track Proceedings}, pages 1--15, 2015.

\bibitem{Loshchilov2017}
Ilya Loshchilov and Frank Hutter.
\newblock {Sgdr: s}.
\newblock pages 1--16, 2017.

\bibitem{Krawczyk2016}
Bartosz Krawczyk.
\newblock {Learning from imbalanced data : open challenges and future
  directions}.
\newblock {\em Progress in Artificial Intelligence}, 5(4):221--232, 2016.

\bibitem{Kolouri2019}
S.~Kolouri, N.~Ketz, X.~Zou, J.~Krichmar, and P.~Pilly.
\newblock {Attention-based selective plasticity}.
\newblock {\em arXiv}, (March), 2019.

\bibitem{Kirichenko2021}
Polina Kirichenko, Mehrdad Farajtabar, Dushyant Rao, Balaji Lakshminarayanan,
  Nir Levine, Ang Li, Huiyi Hu, Andrew~Gordon Wilson, and Razvan Pascanu.
\newblock {Task-agnostic Continual Learning with Hybrid Probabilistic Models}.
\newblock pages 1--23, 2021.

\bibitem{Amendola2020}
Carlos Am{\'{e}}ndola, Alexander Engstr{\"{o}}m, and Christian Haase.
\newblock {Maximum number of modes of Gaussian mixtures}.
\newblock {\em Information and Inference}, 9(3):587--600, 2020.

\end{thebibliography}

}
\end{document}